\documentclass{article}

\usepackage{arxiv}

\usepackage[utf8]{inputenc} 
\usepackage[T1]{fontenc}    
\usepackage{hyperref}       
\usepackage{url}            
\usepackage{booktabs}       
\usepackage{amsfonts}       
\usepackage{nicefrac}       
\usepackage{microtype}      
\usepackage{lipsum}
\usepackage{graphicx}
\usepackage{xcolor}

\usepackage{lineno,hyperref}
\usepackage{epsfig}
\usepackage{amsmath}
\usepackage{amssymb}
\usepackage{subfigure}
\usepackage{float}
\usepackage[export]{adjustbox}
\usepackage{rotating}
\usepackage{tikz}
\usepackage{enumitem}

\modulolinenumbers[5]

\newcommand{\figref}{Fig. \ref}

\title{A Novel Online Action Detection Framework from Untrimmed Video Streams}

\author{
 Da-Hye Yoon \\
  Department of Computer and Radio Communications Engineering\\
  Korea University\\
  Anam-dong, Seongbuk-ku, Seoul 02841, Korea \\
  \texttt{dahye417@korea.ac.kr} \\
   \And
 Nam-Gyu Cho \\
  Department of Brain and Cognitive Engineering\\
  Korea University\\
  Anam-dong, Seongbuk-ku, Seoul 02841, Korea \\
  \texttt{southq@korea.ac.kr} \\
   \And
 Seong-Whan Lee \thanks{Corresponding author}\\
  Department of Artificial Intelligence\\
  Korea University\\
  Anam-dong, Seongbuk-ku, Seoul 02841, Korea \\
  \texttt{sw.lee@korea.ac.kr} \\
}

\begin{document}
\maketitle
\begin{abstract}
Online temporal action localization from an untrimmed video stream is a challenging problem in computer vision. It is challenging because of i) in an untrimmed video stream, more than one action instance may appear, including background scenes, and ii) in online settings, only past and current information is available. Therefore, temporal priors, such as the average action duration of training data, which have been exploited by previous action detection methods, are not suitable for this task because of the high intra-class variation in human actions.
We propose a novel online action detection framework that considers actions as a set of temporally ordered subclasses and leverages a future frame generation network to cope with the limited information issue associated with the problem outlined above. Additionally, we augment our data by varying the lengths of videos to allow the proposed method to learn about the high intra-class variation in human actions. We evaluate our method using two benchmark datasets, THUMOS'14 and ActivityNet, for an online temporal action localization scenario and demonstrate that the performance is comparable to state-of-the-art methods that have been proposed for offline settings.
\end{abstract}

\keywords{Online action detection \and Untrimmed video stream \and Future frame generation \and 3D convolutional neural network \and Long short-term memory}

\label{sec_intro}
Humans are unable to observe a video stream, such as footage streamed from surveillance cameras or a live TV show, continuously without taking any breaks and perform timely decisions based on their observations. For this aspect, online action detection\footnote{In computer vision literature, ``action detection'' and ``temporal action localization'' are used interchangeably. In this paper, we use some related terms as follows. ``Temporal localization'' means finding the starting and ending points of action. ``Classification'' means predicting the label of action. ``Action detection'' means performing temporal localization and action classification simultaneously.} through computer vision plays a vital role in many real-world applications, such as video surveillance systems, pet monitoring systems, elderly (or child) care services, to name a few.
However, the nature of video streams\footnote{We use the term ``video stream'' to indicate continuously incoming data in a stream format (i.e., one frame per moment) and the term ``video'' (or video clip) to indicate data stored on a storage device.} makes this task particularly challenging. Streams are provided in an untrimmed format, meaning multiple actions (or action instances) can occur together with background scenes containing no actions.

Most existing action recognition methods \cite{BLOOM2017, CARMONA2018443, TU201832} focus on trimmed (pre-segmented) video clips that include only a single action. Therefore, applying these methods directly to untrimmed videos is unsuitable for online action detection tasks. In recent years, action recognition from untrimmed videos has been the focus of several challenge competitions, such as THUMOS'14 \cite{THUMOS14} and ActivityNet \cite{caba2015activitynet}. There have been several pioneering studies on action detection tasks, such as Gaidon et al. \cite{gaidon2013temporal}. Xu et al. localize actions once the whole input streams of a video clip are processed
\cite{xu2017r}. \textcolor{black}{Zhao et al. accumulate confidence scores to decide whether a proposal is meaningful, then to classify as a valid action class \cite{zhao2017temporal}}. However, the approaches mentioned above only consider the task in offline settings, which means that detection is performed using video clips.

\begin{figure}
\label{fig_intro}
\centering
\includegraphics[trim=1.4cm 0.5cm 1.3cm 0.7cm, clip=true, width=.6\linewidth ]{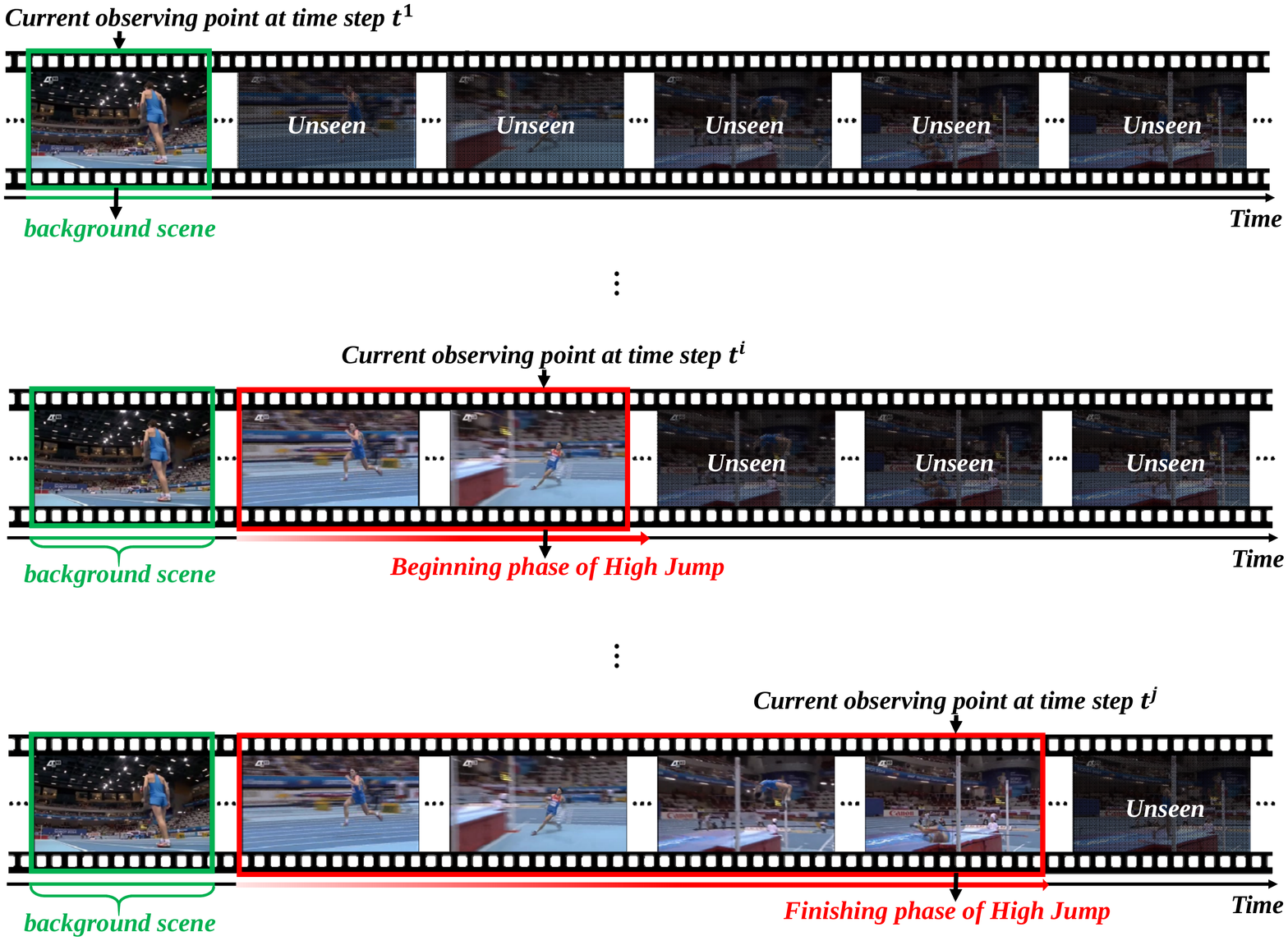}
\caption{Illustration of online action detection. The goal is to determine the action type and its temporal location (start and end points) from a live video stream. Desired output is as follows: (top row) at current time $t^1$ the system determines that the input contains background scene, (middle row) after some time steps ($t^i$) it detects that High Jump action has started, and (bottom row) again after some time steps ($t^j$) it detects that High Jump action has finished. Note that the system has to decide with a limited amount of information.}
\end{figure}
This problem becomes even more challenging in online settings because future observations are not available (see Fig. \ref{fig_intro}). In other words, only past and current information are available for deciding at a given moment. This situation can be compared to the situation where a person is watching a live TV show. He can only know what has happened \textcolor{black}{so far,} and must predict what will happen in the future. In contrast, in an offline setting, he could freely scrub forward or backward in a video clip to obtain information.
Therefore, it is difficult to find the ending point of action in online settings because there is uncertainty as to whether or not an action will continue. To address this issue, Li et al. \cite{li2016online} used temporal priors, such as the average duration of action. However, their performance is not satisfactory because temporal priors do not accurately capture the high intra-class variation of action lengths. To summarize, in order to perform online action detection, the properties of video streams, which come in an untrimmed format with limited information, as well as the high intra-class variation of human actions, must be appropriately handled.

In this paper, we propose a novel online action detection framework (Fig. \ref{fig_frm}). To cope with the limited information problem in online settings, we generate future input frames by leveraging a future frame generation network \cite{villegas2017decomposing}. To address the issue of high intra-class variation, we model each action class as a set of temporally ordered subclasses, called \emph{beginning and finishing phases}, to estimate temporal information directly from input streams with no dependency on prior learning from training data. We exploit a 3D convolutional neural network (3DCNN) structure \cite{tran2015learning} and classify action classes using a sliding window. The efficiency of the sliding window method has been discussed in several studies \cite{shou2016temporal,wang2014action}, which have demonstrated its ability to consider temporal contextual information. Additionally, multi-scale sliding windows tend to perform better than single-scale windows. However, we exploit the single-scale method because multi-scale windows are computationally expensive and require a post-processing operation, such as non-maximum suppression (NMS), which is unsuitable for online settings. To overcome the limitations of single-scale sliding windows, we augment our data by varying the lengths of videos during training. To summarize, the contributions of this study are three-fold:
\begin{itemize}
\item{}{We introduce a future frame generator for online action detection to compensate for the limited information available in online settings.}
\item{}{We decompose each action class into two temporally-ordered subclasses to estimate temporal information directly from visual input.}
\item{}{We augment data by varying the lengths of videos for further improvements of the single-scale sliding window method.}
\end{itemize}

We first review related work in Section 2. In Section 3, we present our framework for online action detection. Section 4 contains our experimental results. 
\section{Related Work}
In this section, we will first review different approaches to action detection. We will then discuss recent works related to our approach, with a focus on \emph{action detection} and \emph{image generation}. \textcolor{black}{Before deep learning, several issues, such as view-independency \cite{Roh2010}, multi-modality \cite{Maeng2012}, and considering human identity \cite{Park2013}, have been challenges for action recognition task. For further information, we refer readers to
\cite{poppe2010survey}.}

\vspace{.3cm}
\noindent\textbf{Action detection}: Action detection is a new field when compared to the action classification task. In recent work action classification, Ma et al. \cite{MA2017334} studied utilizing web images for better model generalization. Ijjina and Chalavadi \cite{IJJINA2016199} exploited a genetic algorithm to train CNN models efficiently. In the early stages of action detection, Hoai et al. \cite{hoai2011joint} performed joint segmentation and recognition on concatenated trimmed videos. Gaidon et al. \cite{gaidon2013temporal} localized actions temporally within the Coffee and Cigarettes dataset, which contains untrimmed video with two classes. Since then, Shou et al. \cite{shou2016temporal} have studied for action detection on untrimmed video datasets, such as THUMOS'14 \cite{THUMOS14} and ActivityNet \cite{caba2015activitynet}.
With excellent results in the THUOMS'14 competition, Oneata et al. \cite{oneata2014lear} and Wang et al. \cite{wang2014action} employed improved dense trajectories (IDT) encoded by fisher vectors (FVs) and CNN features, as well as SVM classifiers based on sliding windows with some variations, such as window sizes, fusion methods, post-processing methods. Inspired by these works, Shou et al. \cite{shou2017cdc} proposed multi-stage CNNs that consider spatial-temporal information and a convolution-de-convolutional (CDC) network that performs spatial downsampling and temporal upsampling to capture abstractions for action semantics and temporal dynamics. Yuan et al. \cite{yuan2016temporal} captured temporal context by proposing a pyramid of score distribution features (PSDF) descriptor. Yeung et al. \cite{yeung2016end} proposed a frame-wise end-to-end framework with a reinforced learning.
Modeling actions in a grammatical form through N-grams and latent Dirichlet allocation (LDA) was performed as another method of action detection \cite{richard2016temporal}. 

Meanwhile, some works considered action detection not only temporally but also spatially. Lan et al. \cite{lan2015action} and Weinzaepfel et al. \cite{weinzaepfel2015learning} proposed a spatio-temporal action localization (or parsing) method by representing mid-level action elements using a hierarchical structure and localizing actions with a spatio-temporal motion histogram (STMH) descriptor individually at the track level. Additionally, Lea et al. \cite{lea2016segmental} detected fine-grained actions in ego-centric video datasets using graphical models based on tracking hands and objects or using deep neural networks.

While most of the methods mentioned above focus on offline settings, several methods performed online action detection by predicting an action's ending point using temporal priors or unsupervised learning. Recently, new datasets for online action detection have been proposed. Geest et al. \cite{de2016online} introduced an RGB-based \emph{TVSeries} dataset with baseline models using long short-term memory (LSTM) and a CNN. Li et al. \cite{li2016online} introduced a skeleton-based online action dataset (OAD) and proposed a joint classification regression method using LSTM.

\vspace{.3cm}
\noindent\textbf{Temporal sliding window-based detection}: It is crucial to consider the temporal contextual information of time series data, such as video and sound. In action detection literature, many methods extensively employed the temporal sliding window method, which is performed by moving a window of a specific size over time. To make the most of temporal contextual information, \cite{shou2016temporal, yuan2016temporal} used multi-scale sliding windows. Features are extracted using windows of various scales (e.g., 16, 32, or 64 frames),  and the detection results of each scale are post-processed to generate a final prediction. However, this approach is unsuitable for online action detection because only a limited amount of information is available. In this paper, despite the success achieved through the use of multi-scale windows, we employ the single-scale window approach.

\vspace{.3cm}
\noindent\textbf{Image generation}: Several approaches have been proposed for generating target images from given images. Kingma et al. \cite{kingma2013auto} proposed a variational inference-based method by extending the auto-encoder structure. Dosovitskiy et al. \cite{7469347} introduced a method applying CNN structure for object-oriented image generation. Additionally, Radford et al.  \cite{radford2015unsupervised} exploited generative adversarial networks (GANs) \cite{goodfellow2014generative}, which reconstruct distributions of data through discriminate networks and adversarial networks.
Based on these approaches, several video frame generation approaches have been proposed. Finn et al. \cite{finn2016unsupervised} proposed a method that generates future frames through a combination of RNNs and auto-encoding (called a variational auto-encoder). Mathieu et al. \cite{mathieu2015deep} and Vondrick et al. \cite{vondrick2016generating} proposed video frame generation methods exploiting deep convolutional GANs \cite{radford2015unsupervised}, respectively. \textcolor{black}{Most recently, Villegas et al. \cite{villegas2017decomposing} proposed a motion-content network (MCnet) that considers temporal and spatial information separately, which resembles the way human brain processes temporal and spatial information \cite{Bulthoff2002}}. In this paper, in order to resolve the limited information issue in online settings, we adopt MCnet for generating future frames.
\section{Method}
In this section, we first overview the proposed framework. Then, we detail our proposed framework, including the architectures of the deep neural networks used in each component. We also elaborate on how to train the networks on a large-scale dataset.

\begin{figure*}[t!]
\centering
\includegraphics[trim=1cm 1.5cm 1cm 1.5cm, clip=true, width=.7\linewidth]{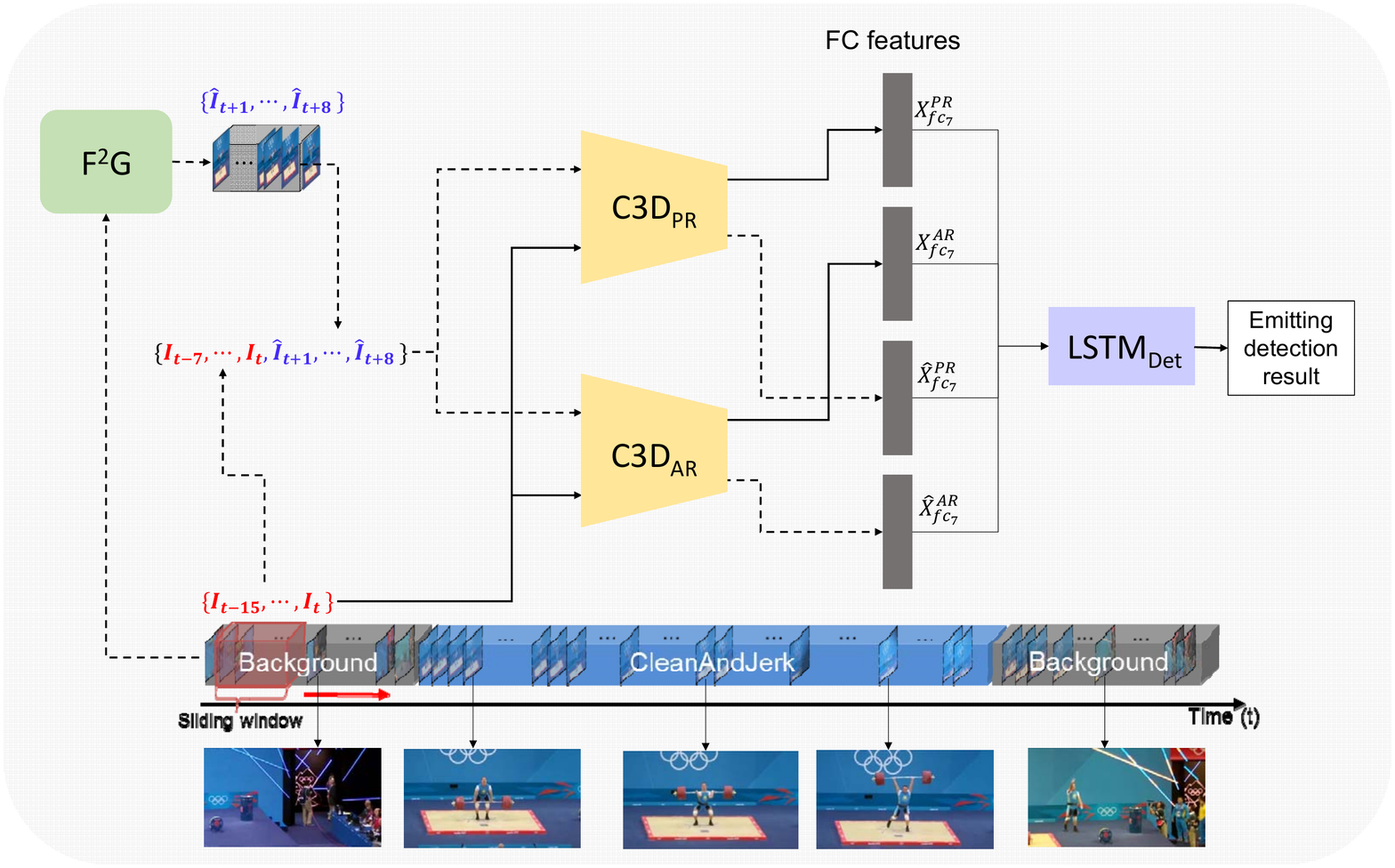}
\caption{Overview of the proposed framework. At the current time $t$, input frames ($I_{t-15}, ..., I_t$) are passed to both PR network (C3D$_{\textrm{PR}}$) and AR network (C3D$_{\textrm{AR}}$) to extract features $X^{PR}_{fc7}$ and $X^{AR}_{fc7}$ from fully connected layer 7 (fc$_7$) of each network. The input frames are also fed into F$^2$G network, which predicts future frames ($I_{t+1}, ..., I_{t+8}$). Then new input frames ($I_{t-7}, ..., I_{t+8}$) are also passed to PR and AR networks to extract features $\hat{X}^{PR}_{fc7}$ and $\hat{X}^{AR}_{fc7}$. All four features are concatenated and passed to the detection network, which emits the detection results by considering both current and (generated) future input frames.}
\label{fig_frm}
\end{figure*}

The objective of our framework is to detect actions in untrimmed video streams. The framework is composed of four deep networks: a proposal representation (PR) network that discriminates between actions and background scenes (Sec. \ref{sec_proposal}), action representation (AR) network that predicts the type and temporal order of an action (Sec. \ref{sec_classify}), future frame generation (F$^2$G) network that generates future video frames (Sec. \ref{sec_f2g}), and detection network that detects actions by receiving outputs from other networks (Sec. \ref{sec_det}). Fig. \ref{fig_frm} illustrates the pipeline of the proposed framework.

Motivations for choosing these networks are as follows. Unlike the sole action classification task, action detection from untrimmed videos requires action representations dedicated to not only action itself but also background scenes. Intuitively, visual treats of background scenes and actions are different. Thus, in the proposed framework, we exploit two deep networks to solve two different tasks: one is to distinguish background scenes from actions, and the other is to classify actions of interest, e.g., twenty classes for THUMOS'14. Both networks have the same structures, 3D convolutional layers followed by fully connected layers (Fig. \ref{fig_proposal}), which have shown outstanding performance for the action classification. When it comes to online situations, as described in Sec. 1, the action localization task suffers from the short of information to solve the problem. In order to resolve this issue, we propose using the future frame generation network. The detection network is composed of LSTM layers to mode temporal correlations and capture temporal changes locally, such as motion features. 

\subsection{Detecting Action Candidate Spots}
\label{sec_proposal}
Untrimmed videos are an irregular combination of actions and background scenes. Under this situation, it is necessary that detecting candidate segments where an action is likely to occur, i.e., distinguishing between actions and background scenes. Towards this goal, we train a proposal representation (PR) network. The PR network takes a video segment as an input, which is acquired by a temporal sliding window of length $\tau$. The network is trained to classify binary classes -- \emph{background scene class and action scene class} -- via a 3DCNN whose final fully connected layer has two neurons.

\begin{figure}
\centering
\includegraphics[trim=1.5cm 8.4cm 1.5cm 8.4cm, clip=true, width=.8\linewidth]{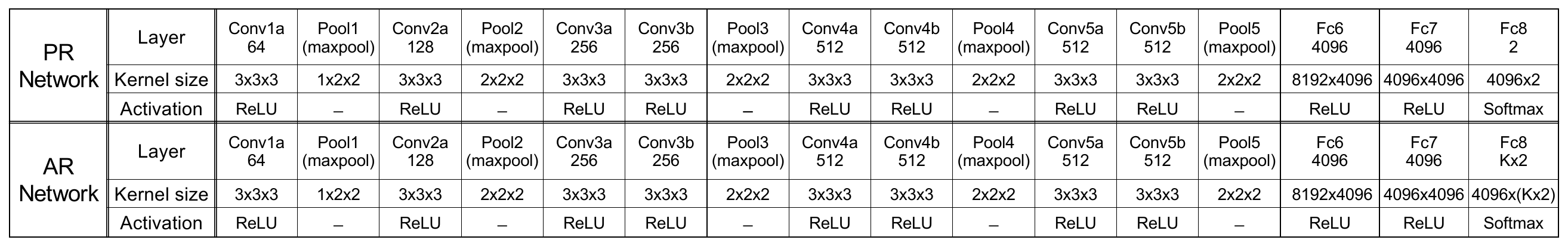}
\caption{The 3DCNN architecture used in the PR and AR networks.}
\label{fig_proposal}
\end{figure}

\vspace{.3cm}
\noindent\textbf{Network architecture}: We employ a 3DCNN to consider spatial and temporal information simultaneously. Different from widely used conventional 2D CNNs, our 3DCNN learns motion context, which is an important clue in video analysis, by considering adding a time axis to the 2D image coordinate system. We adopt the 3DCNN architecture \cite{tran2015learning}: 8 convolutional layers, 5 pooling layers, and 2 fully connected layers, and an output layer with a softmax function. Details of the architecture are shown in Fig. \ref{fig_proposal} (top row).

\subsection{Learning Visual Traits from a Temporal Order}
\label{sec_classify}
\begin{figure}
\centering
\includegraphics[width=.5\linewidth]{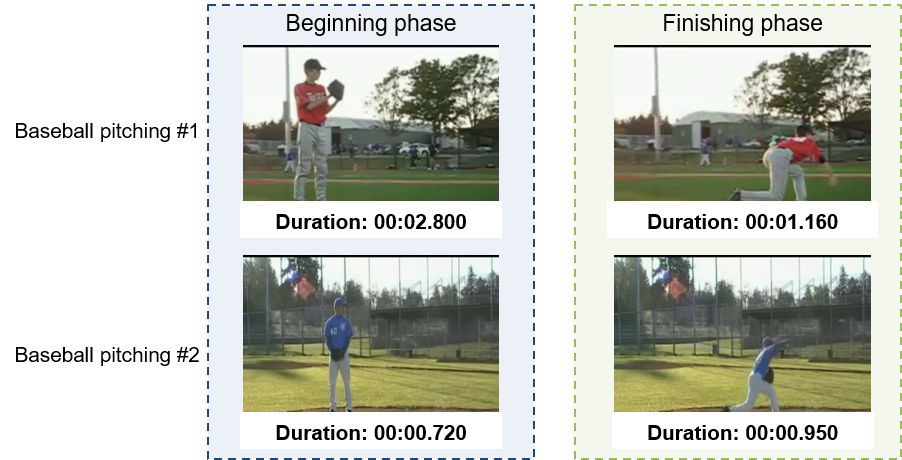}
\caption{Visualization of each phase of baseball pitching.}
\label{fig_cls_ex}
\end{figure}

The beginning and ending phases of all action instances in the same action class share an identical trait. E.g., assuming there are videos containing scenes where a person is throwing a baseball, the beginning phase of all videos would contain the person making a wind-up posture before throwing the ball, and the ending phase would contain the person leaning forward and lifting his leg back after throwing the ball. However, duration of these phases are different per instance (see \figref{fig_cls_ex}). In order to capture this trait, we design an action representation (AR) network considering actions as a set of temporally ordered subclasses. Specifically, we divide each action class into the beginning and ending phase classes, and train a 3DCNN to classify these subclasses.

Learning temporally ordered subclasses allows the model to represent the time phase of an action using only visual information. When compared to methods that exploit the average length of each action as temporal priors to detect actions, e.g.,  \cite{li2016online}, our method detects only from a given input sequence.

\vspace{.3cm}
\noindent\textbf{Network architecture}: The AR network has the architecture identical to the PR network, except the last fully connected layer consists of $K \times 2$ neurons where $K$ is the number target classes, e.g., $K=20$ for THUMOS'14 and $K=200$ for ActivityNet. Details of the architecture of the AR network are shown in Fig. \ref{fig_proposal} (bottom row).

\subsection{Generating Future Frames}
\label{sec_f2g}
The major limitation in online settings is that only past and current information can be considered for decision. To overcome this limitation, we introduce a future frame generation (F$^2$G) network that generates future frames. In this paper, we design an F$^2$G network with the same architecture as the MCnet proposed in \cite{villegas2017decomposing}, which considers spatial and temporal information by modeling content and encoding motion, respectively.

\begin{figure}
\centering
\includegraphics[trim=0cm 0cm 0cm 0cm, clip=true, width=.6\linewidth]{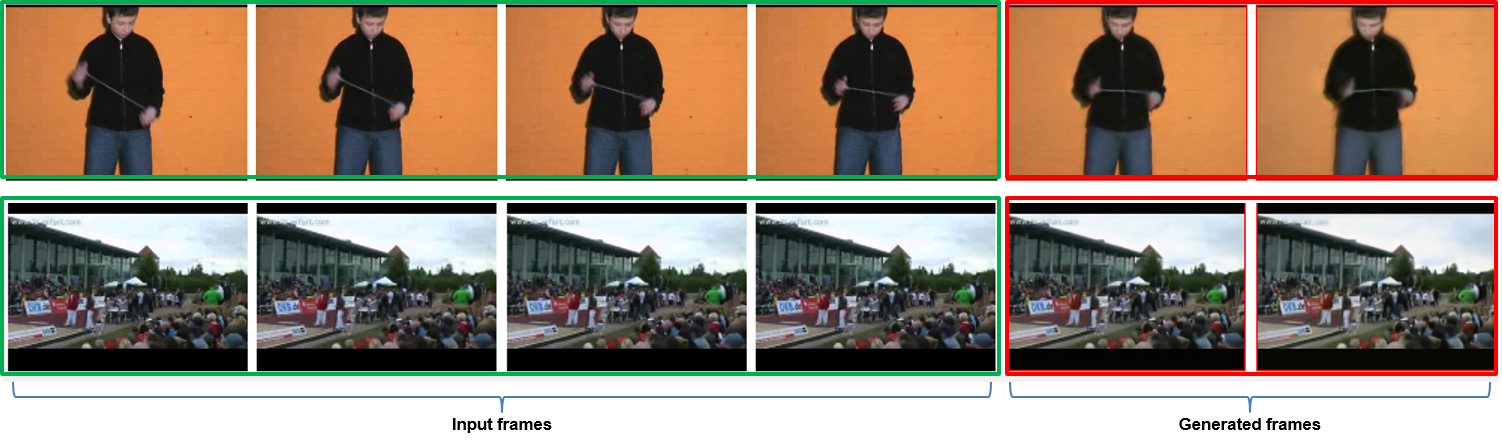}
\caption{Illustration of future frame generation. (From left to right) the first four frames with green bounding boxes are input frames, and the next two frames with red bounding boxes are generated frames.}
\label{fig_f2g_ex}
\end{figure}

\vspace{.3cm}
\noindent\textbf{Network architecture}: The F$^2$G network is an encoder-decoder network composed of two different networks: a content encoder with a CNN architecture and motion encoder with a convolutional LSTM architecture. Generated samples are illustrated in \figref{fig_f2g_ex}. We refer to \cite{villegas2017decomposing} for further details.

\subsection{Detecting Actions by Modeling Temporal Correlations}
\label{sec_det}
\begin{figure}
\centering
\includegraphics[width=.35\linewidth]{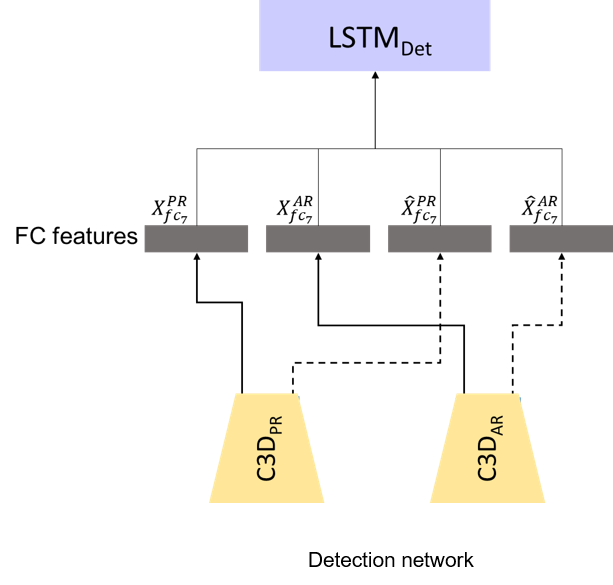}
\caption{Illustration of detection network. The outputs of final fully connected layers (not softmax layer) of C3D$_{PR}$ and C3D$_{AR}$ are concatenated and then fed to the LSTM network to detect action.}
\label{fig_det_ex}
\end{figure}

To detect actions, to model temporal correlations and to capture temporal changes locally, such as motion features, are an essential factor. However, the PR and AR networks employing 3DCNN are lack of this ability. Therefore, we design our detection network by employing a recurrent neural network (RNN) that can model temporal correlations. The network takes the outputs from each fully connected layer (fc$_7$) of the PR and AR networks as input (see Fig. \ref{fig_det_ex}). The detection network uses the outputs of other networks to reflect the response (opinion) of each network (expert) for a given input data sample over time; it then derives final results by modeling temporal correlations from each network using the RNN.

\begin{figure}
\centering
\includegraphics[trim=14.15cm 9.45cm 7cm 9.05cm, clip=true, width=.32\linewidth]{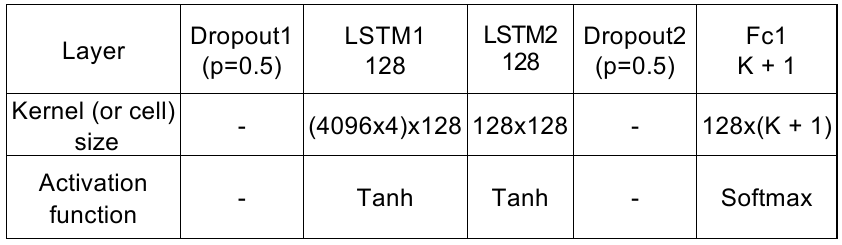}
\caption{The LSTM architecture used in the detection network.}
\label{fig_detection}
\end{figure}

\vspace{.3cm}
\noindent\textbf{Network architecture}: There are various types of RNNs, such as long short-term memory (LSTM) and gated recurrent units (GRUs). In this paper, the detection network consists of a dropout layer with the probability 0.5, two LSTM layers, each of which has 128 states, a dropout layer with the probability 0.5, and a fully connected layer having $K + 1$ neurons, which correspond to $K$ action classes and one background class of a target dataset, e.g., $K = 20$ for THUMOS'14 and $K=200$ for ActivigyNet. Fig. \ref{fig_detection} shows details of the architecture of the detection network.

\subsection{Training}
\label{sec_train}
\noindent\textbf{PR and AR networks}:
For training PR and AR networks, we first initialize the weights of all convolution layers (Conv1a to Conv5b) and the first fully connected layer (fc6) with the pre-trained 3DCNN network \cite{tran2015learning}. Then we fine-tune these networks with the target benchmark dataset, either THUMOS'14 or ActivityNet. To train the PR network, we modified labels as two: foreground (action) and background (non-action). To train the AR network, we divide each action class into two subclasses, beginning and ending, thus 40 classes for THUMOS'14 and 400 classes for ActivityNet.

In experiments, we use SGD (stochastic gradient descent) optimization with a learning rate of 0.0001, momentum of 0.9, weight decay factor of 0.0005, and dropout probability of 0.5. We use the cross-entropy as the loss function to update the network weights.

\begin{equation}
\mathcal{L}_{softmax}=\frac{1}{N}\sum_{n}^{N}\sum_{k}^{K}
-y^{k}log(\hat{y}^k),
\end{equation}
\noindent where $N$ is the total number of training samples, $y_{n}^k$ is the $k$-th element value of the ground truth probability of the $n$-th sample. $\hat{y}_{n}^k$ is the $k$-th element value of the predicted probability (network output) of the $n$-th sample. $K$ is the total number of classes: 2 for the PR network and the number of action classes of a target dataset for the AR network.

\vspace{.3cm}
\noindent\textbf{F$^2$G network}: We fine-tune the F$^2$G network on the action classes of a target dataset. The F$^2$G network uses the loss function composed of different sub-losses: an image loss and generator loss (we refer to \cite{villegas2017decomposing} for further details). In experiments, we fine-tune the network with a learning rate of 0.001.

\vspace{.3cm}
\noindent\textbf{Detection network}:
We use SGD (stochastic gradient descent) optimization with a learning rate of 0.0001, momentum of 0.9, weight decay factor of 0.0005, and dropout probability of 0.5. We use the cross-entropy as the loss function to update the network weights. Additionally, we use classes weight to consider imbalanced instance number among classes as follows:

\begin{equation}
\omega_{k} = 1 - \frac{|S_k|}{2\hat{|S|}}
\end{equation}
\noindent where $|S_k|$ is the number of training instances of $k$-th class and $|\hat{S}|$ is the largest instance number among all classes.
In experiments, we use RMSProp optimization with a learning rate of 0.0001 and a dropout probability of 0.5. We use the cross-entropy as the loss function.
\subsection{Data Augmentation}
\label{sec_dataaug}
We introduce a video data augmentation technique for further improvement of the single-scale temporal sliding window method. There are some issues in both the single-scale window and multi-scale window methods. When using single-scale windows, they cannot capture areas that are important for representing an action if the length of the window is set improperly. When using multi-scale windows, they require post-processing techniques that are unsuitable for online settings. Therefore, we augment training data by varying the lengths of videos. This augmentation allows us to obtain effects similar to using multi-scale windows even though we only use single-scale windows. 

\begin{figure}
\centering
\includegraphics[trim=1cm 7cm 1cm 6.5cm, clip=true, width=.6\linewidth]{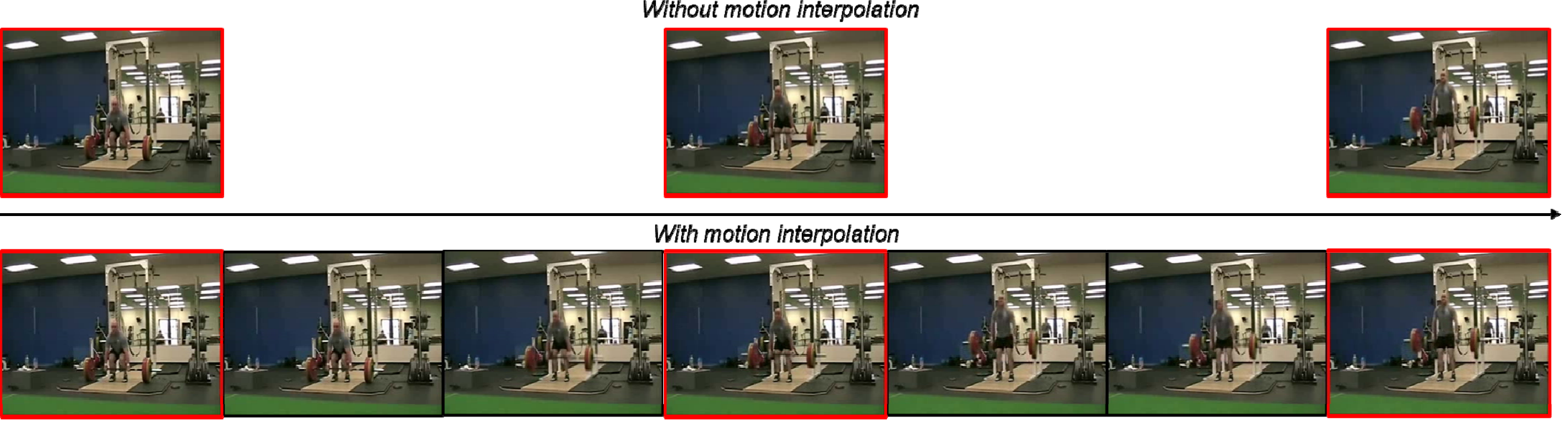}
\caption{Illustration of motion interpolation.}
\label{fig_motion_interpol}
\end{figure}

We conduct augmentation in two ways: increasing and decreasing. The former is to simulate a video clip being played faster by sampling some frames from the original video. The later is to simulate a video clip being played slower. This effect is performed by motion interpolation using the butterflow algorithm\footnote{More details for the butterflow algorithm, and its implementation can be found at https://github.com/dthpham/butterflow}. Motion interpolation is performed by rendering intermediate frames between two frames based on motion. Specifically, given two frames $I_A$ and $I_B$, this technique fills the space between these two frames with generated intermediate frames $I_{AB_1}$, $I_{AB_2}$, ..., $I_{AB_O}$, as shown in Fig. \ref{fig_motion_interpol}. 

In general, data augmentation helps a model to better generalize \cite{Krizhevsky_imagenet}. With the proposed data augmentation, a model sees training videos with different temporal resolutions. This augmentation mimics, to a certain extent, the effect when using a multi-scale window. As a multi-scale window would learn from having windows of different temporal scales as input during training time, with the augmented data, a single-scale window learns from temporal variations during training time in a different way.

\section{Experiments}
We evaluated the proposed framework on two large benchmark datasets: THUMOS'14 \cite{THUMOS14} and ActivityNet \cite{caba2015activitynet}. We will first describe the experimental setup, compare the performance with other methods on two benchmark datasets, and then demonstrate the ablation study. Finally, we describe the limitations of the proposed framework.

\subsection{Experimental Setup}
\noindent\textbf{Implementation details}: The length of the temporal sliding window $\tau$ is 16. During the training phase, there is no overlap between sliding windows. In other words, sliding windows at time $t_1$ and $t_2$ do not have an intersection between them. In the test phase, we allow 50\% overlap between adjacent sliding windows. At time $t$, the PR and AR networks take 16 frames ($I_{t-15}, ..., I_{t}$) as an input. The F$^2$G network generates 8 future frames ($\hat{I}_{t+1}, ..., \hat{I}_{t+8}$) from the input frames. The generated 8 frames are concatenated with past input frames ($I_{t-7}, ..., I_t, \hat{I}_{t+1}, ..., \hat{I}_{t+8}$). Then these concatenated 16 frames are passed to the PR and AR networks (Fig. \ref{fig_frm}). The detection network receives a $4096\times4$ dimensional vector as an input which corresponds to the output of the PR network (4096$\times$2, 4096 from the input frames and 4096 from the generated frames concatenated with the second half of input frames) and AR network (4096$\times$2, same as the PR network).

The PR and AR networks were trained with 50 epochs on each training and validation sets of THUMOS'14. The F$^2$G network was trained with 150k epochs on the training set. Training the detection network is a nontrivial part. We first train the network with 100 epochs on the training set, then 100 epochs on the validation set, and we repeat three times. The training batch is organized as 8 time steps, i.e., 8 sliding windows to allow the network to learn the long-term temporal relationship of input frames. In the test phase, a single sliding window is passed to the detection network. During training, we augment the data by increasing the speed by two and decreasing by half.

\vspace{.3cm}
\noindent\textbf{Evaluation metric}: We use interpolated average precision (AP) and mean average precision (mAP) to evaluate the performance of our model following the guidelines of the action detection task of THUMOS'14. A detection result is evaluated as a true positive when the overlapping intersection over union (IoU) between the predicted temporal range ${R}_{pred}$ and ground truth temporal range ${R}_{gt}$ is larger than an overlap threshold $\theta$.

\begin{equation}
IoU =\frac{{R}_{pred}\bigcap {R}_{gt}}{{R}_{pred}\bigcup {R}_{gt}}.
\label{eq_map}
\end{equation}

\subsection{Experimental Results on THMOS'14}
\noindent\textbf{Dataset}: We consider twenty classes of THUMOS'14 dataset for evaluation: BaseballPitch, BasketballDunk, Billiards, CleanAndJerk, CliffDiving, CricketBowling, CricketShot, Diving, FrisbeeCatch, GolfSwing, HammerThrow, HighJump, Javelin-Throw, LongJump, PoleVault, Shotput, SoccerPenalty, TennisSwing, Throw-Discus, and VolleyballSpiking. The training set consists of 2,765 trimmed videos that contain one action. The background set consists of 2,500 untrimmed videos that include actions in undefined categories. The validation and test sets consist of 1,010 untrimmed videos and 1,574 untrimmed videos, respectively, which contain more than one action instance, including backgrounds.

\begin{figure}
\centering
  \begin{adjustbox}{addcode={\begin{minipage}{\width}}{\caption{%
      Samples for 20 classes in THUMOS'14 dataset.
      }\end{minipage}}}
      \includegraphics[width=.6\linewidth]{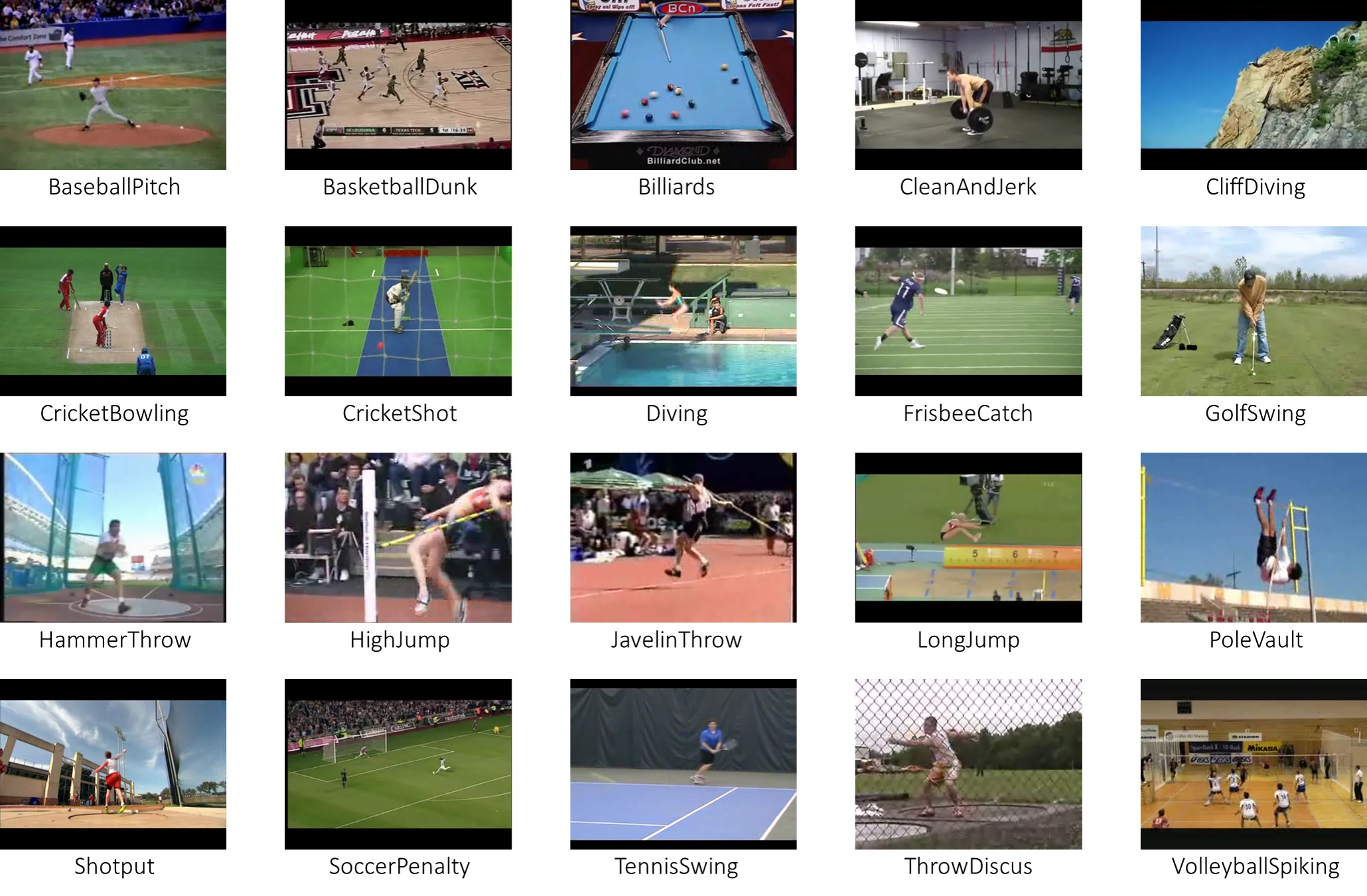}%
  \end{adjustbox}
\label{fig_prsoposal}
\end{figure}

\vspace{.3cm}
\noindent\textbf{Comparison to other methods}: We evaluate our model and compare its performance to various offline action detection methods: Wang et al. \cite{wang2014action}, Oneta et al. \cite{oneata2014lear}, Yeung et al. \cite{yeung2016end}, Richard et al. \cite{richard2016temporal}, Shou et al. \cite{shou2017cdc}, and Zhao et al. \cite{zhao2017temporal} using THUMOS'14 because there are no online methods reported on this dataset. As a baseline, we implemented a framework identical to the proposed method except F$^2$G is removed and trained without data augmentation. We report the performance with mAP metric (\ref{eq_map}) with the threshold $\theta$ ranging from 0.1 to 0.7.

\begin{table}
\caption{Performance of methods on THUMOS'14.}
\centering
\includegraphics[trim=5.3cm 6.3cm 5.3cm 6.3cm, clip=true, width=0.5\linewidth]{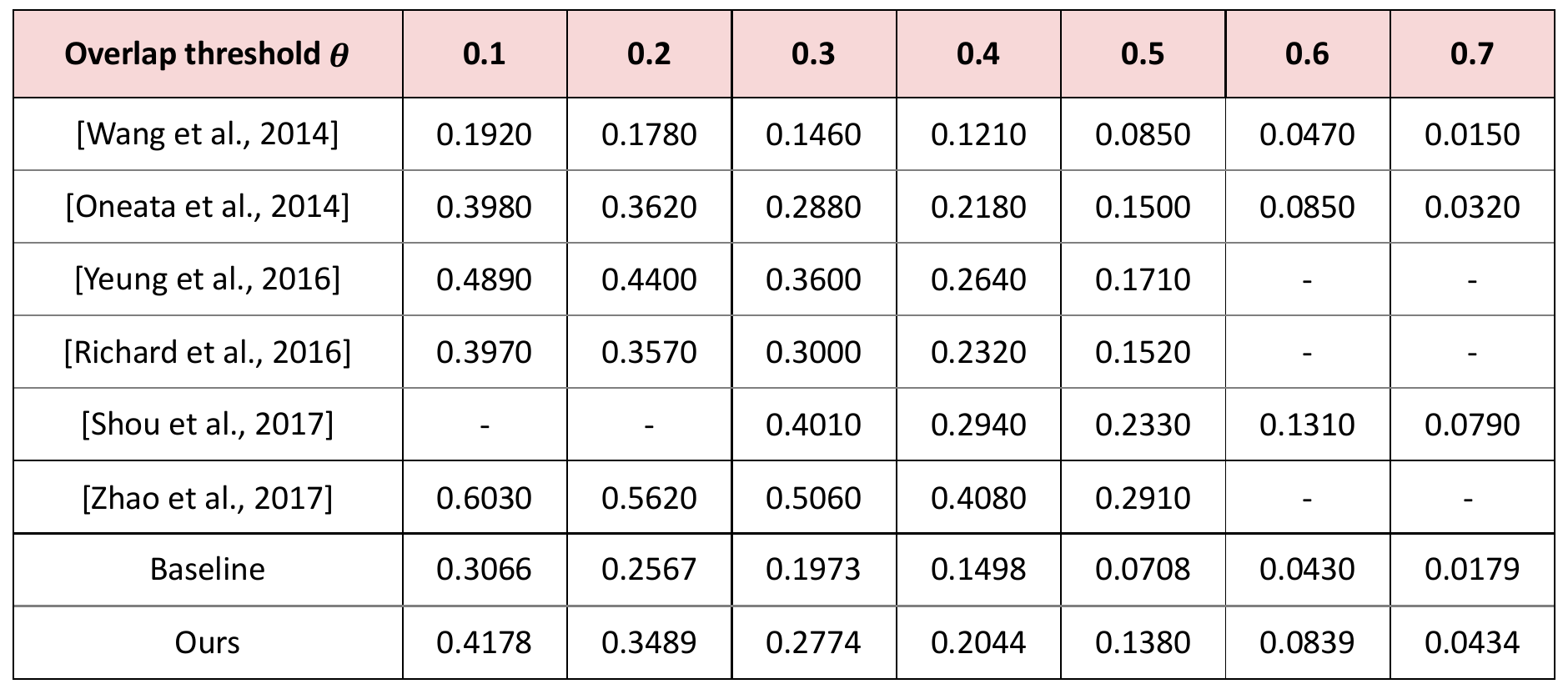}
\label{cmp_table2}
\end{table}

\begin{figure}
\centering
\subfigure[CleanAndJerk]{
  \label{ret_tu1}
  \includegraphics[trim=1.6cm 2.9cm 1.6cm 2.9cm, clip=true, width=0.45\linewidth]{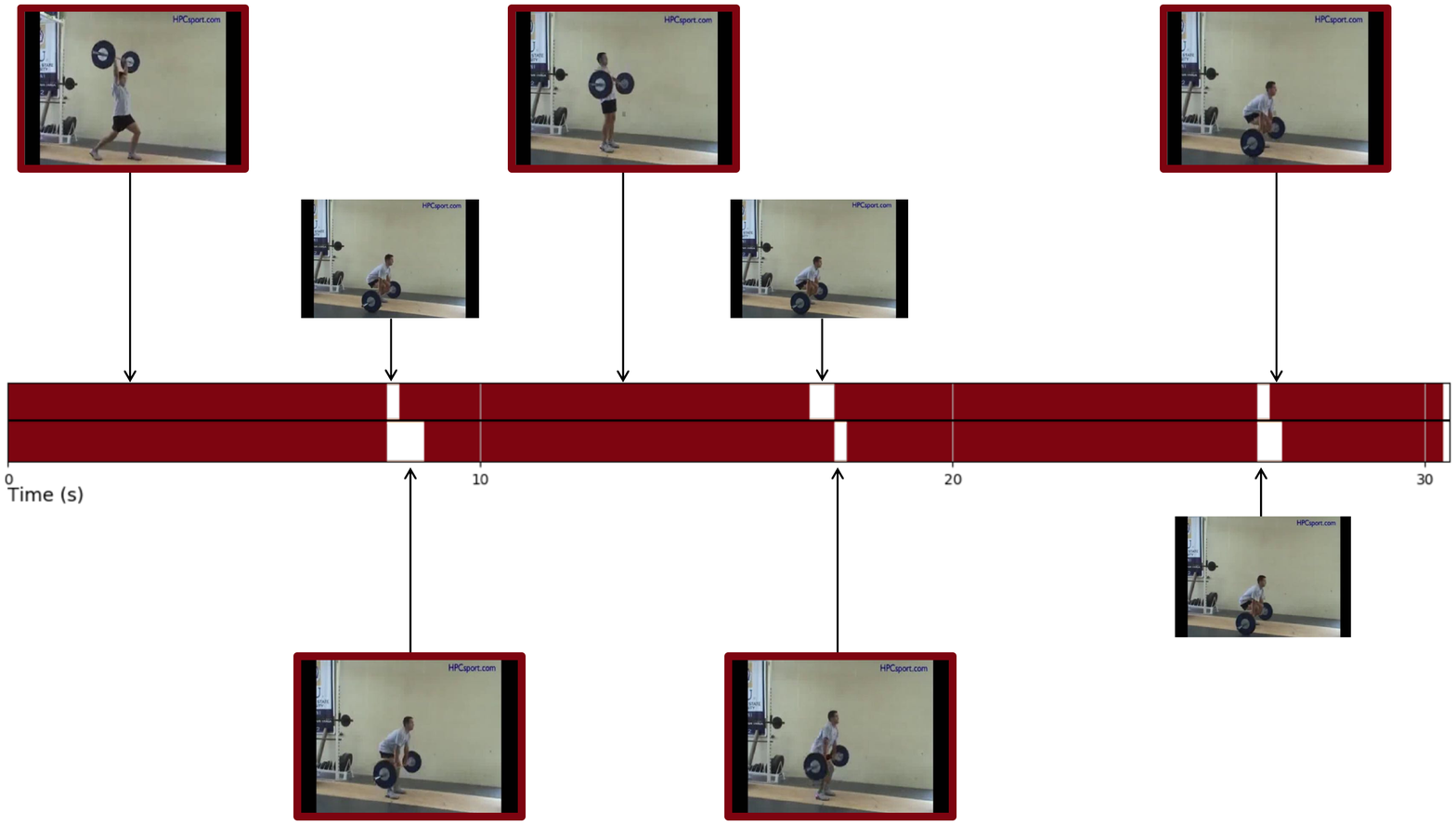}
  }
\subfigure[BasketballDunk]{
  \label{ret_tu2}
  \includegraphics[trim=1.5cm 3.0cm 1.6cm 3.0cm, clip=true, width=0.45\linewidth]{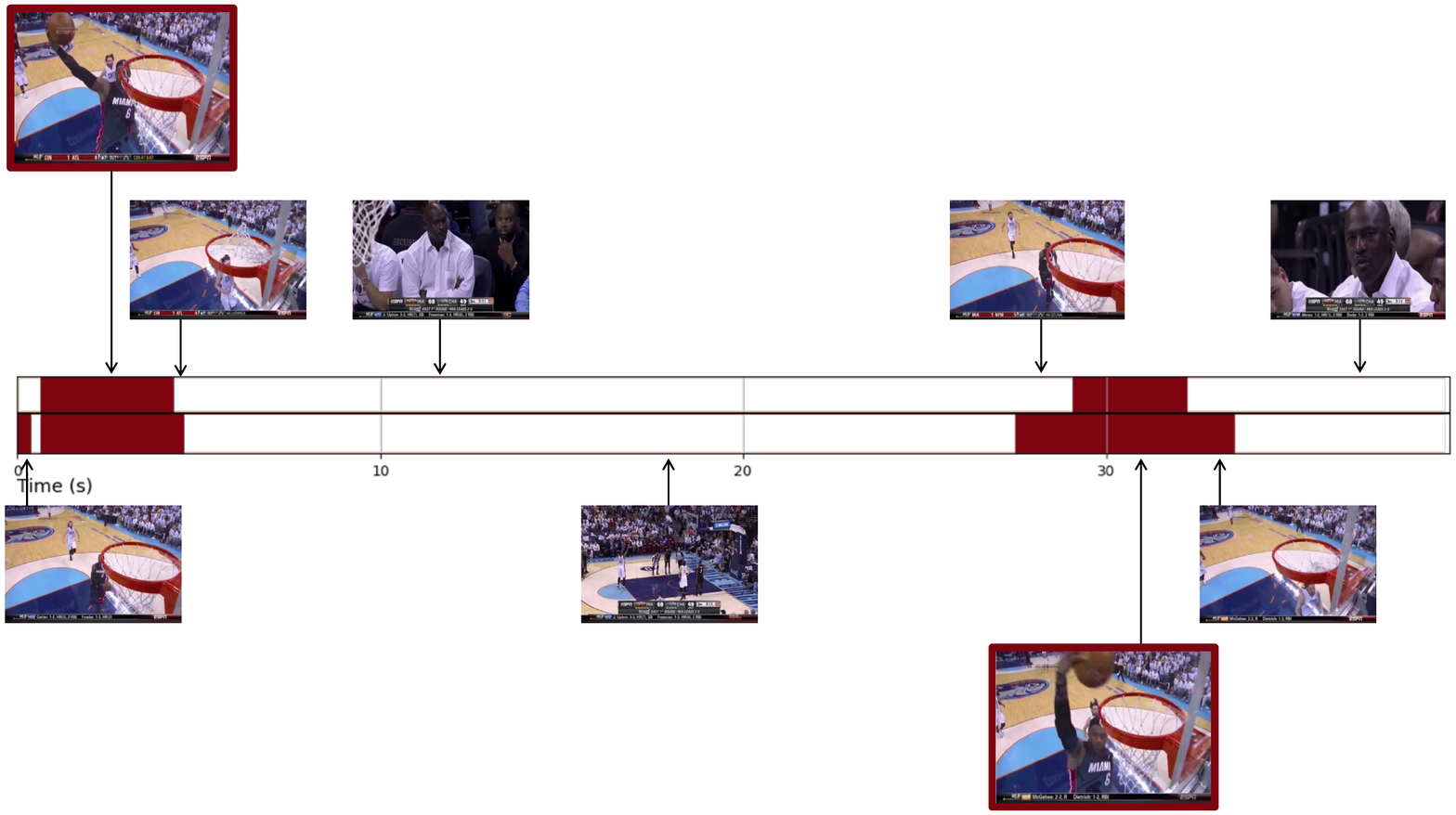}
  }
\caption{\textcolor{black}{Detection results illustration on \emph{CleanAndJerk} (a) and \emph{BasketballDunk} (b) classes of THUMOS'14 test set. In the horizontal bar of each subfigure, the top row denotes the ground truth and the bottom row detection result. The color-shaded region (dark red) indicates the target class and the white region background. Images with the dark red boundary denote the class instance and those without the colored border background (best seen in color). In (a), paused moments of the person sitting down are labeled as background while the rest is CleanAndJerk in which the proposed method detects with some misalignment. In (b), the proposed method struggles to locate the background precisely. It is probably due to short of the temporal context available to decide.}}
\label{fig_result3}
\end{figure}

Tab. \ref{cmp_table2} summarizes the performance of methods on THUMOS'14. Note that the proposed method was tested on the online setting scenario in which only past and current information are available for decision at a given moment, while other comparing methods were tested on the offline setting in which rich temporal context information is available. Taking this into consideration, the performance of our method is comparable to other offline setting methods. Compared to the baseline, the proposed method outperforms significantly, 0.06 higher mAP on average. Fig. \ref{fig_result3} visualizes results in some challenging test data instances on THUMOS'14.

\subsection{Experimental Results on ActivityNet}
\noindent\textbf{Dataset}: We use the latest version 1.3 of ActivityNet dataset, which contains two hundred activity classes. The training set consists of 10,024 untrimmed videos that contain single activity instances. The validation set and test set consist of 4,926 and 5,044 untrimmed videos, respectively. In this experiment, we use the validation set for evaluation since the annotation of the test set is not available.

\vspace{.3cm}
\noindent\textbf{Comparison to other methods}: We evaluate our model and compare its performance to two offline action detection methods: Heilbron et al. \cite{Heilbron_2017_CVPR}, Shou et al. \cite{shou2017cdc}, and Zhao et al. \cite{zhao2017temporal} using the ActivityNet because there are no online methods reported on this dataset. As a baseline, we implemented a framework identical to the proposed method except F$^2$G is removed and trained without data augmentation. We report the performance with mAP metric (3) with the threshold $\theta$s: 0.5, 0.75, and 0.95 as in \cite{Heilbron_2017_CVPR}.

\begin{table}
\caption{Performance of methods on ActivityNet.}
\centering
\includegraphics[trim=9.4cm 7.7cm 9.4cm 7.65cm, clip=true, width=0.3\linewidth]{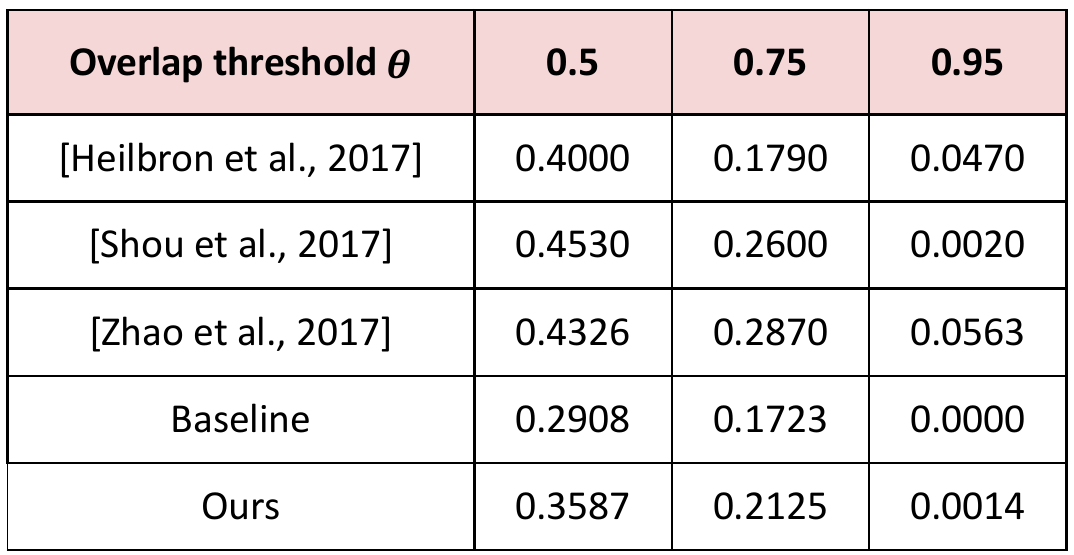}
\label{cmp_table_act}
\end{table}

\begin{figure}
\centering
\subfigure[Knitting]{
  \label{ret_act1}
  \includegraphics[trim=1.5cm 3.0cm 1.55cm 3.0cm, clip=true, width=0.45\linewidth]{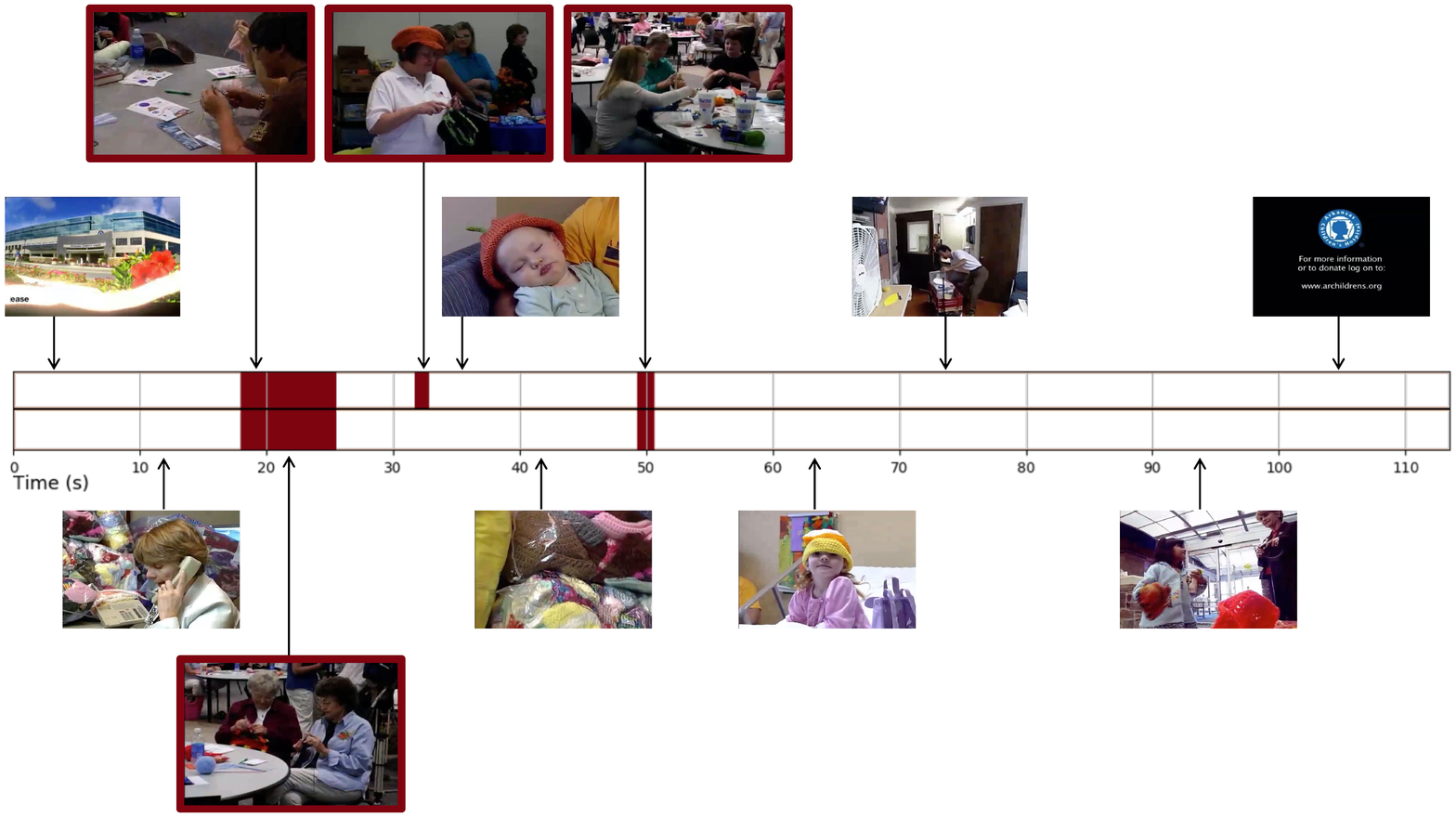}
  }
\subfigure[Tango]{
  \label{ret_act2}
  \includegraphics[trim=1.65cm 2.9cm 1.45cm 2.9cm, clip=true, width=0.45\linewidth]{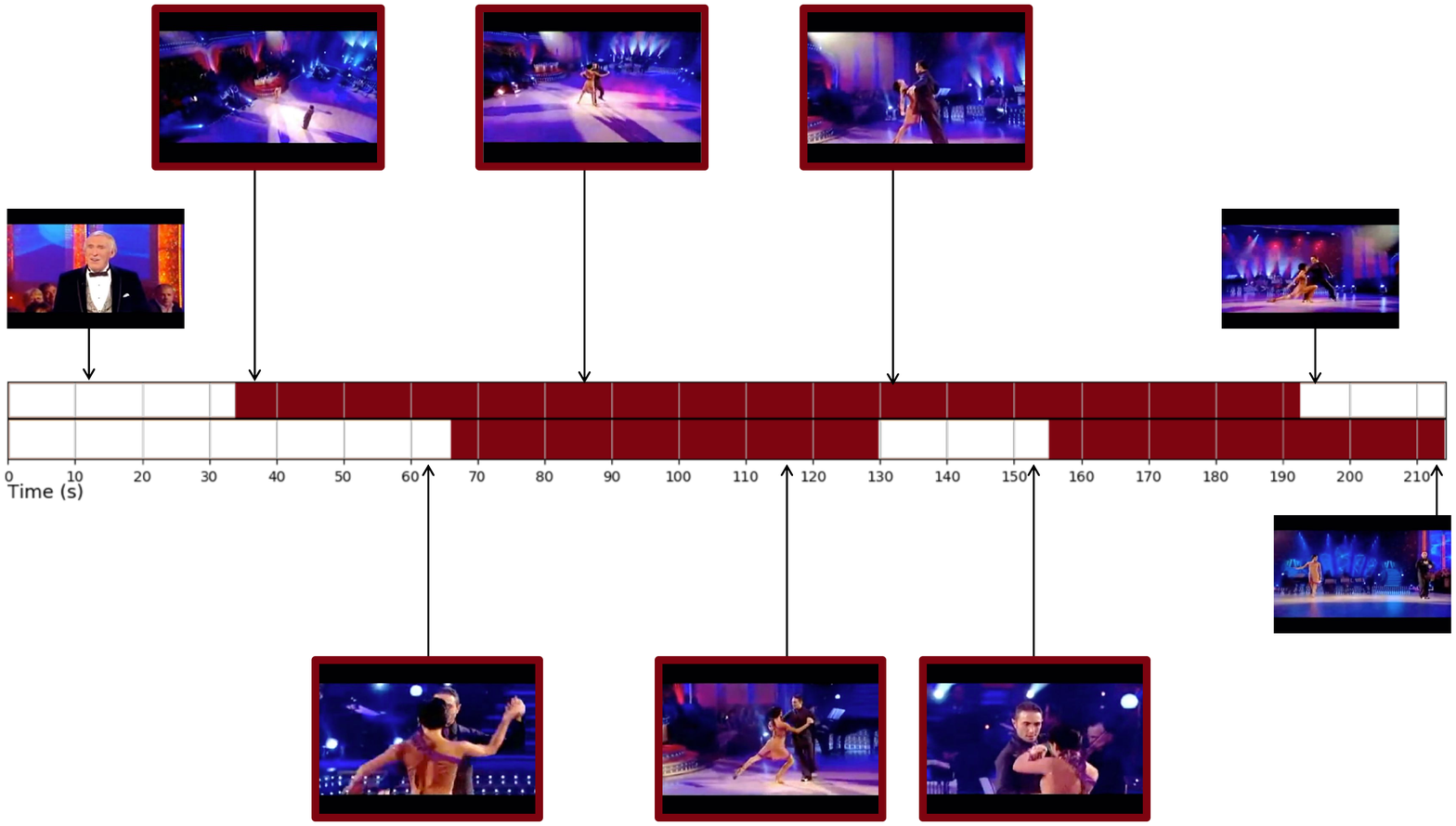}
  }
\caption{\textcolor{black}{Detection results illustration on \emph{Knitting} (a) and \emph{Tango} (b) classes of ActivityNet validation set. In the horizontal bar of each subfigure, the top row denotes the ground truth and the bottom row detection result. The color-shaded region (dark red) indicates the target class and the white region background. Images with the dark red boundary denote the class instance and those without the colored border background (best seen in color). In (a), the proposed method successfully detects two class instances except for one instance of a short period. (b) illustrates the failure case: the proposed method does not locate the instances precisely. We postulate that it is due to the long sequence length of the input stream; thus the proposed method struggles to exploit temporal information.}}
\label{fig_result4}
\end{figure}

Tab. \ref{cmp_table_act} summarizes the performance of methods on ActivityNet. Note that the proposed method was tested on the online setting scenario in which only past and current information are available for decision at a given moment, while other comparing methods were tested on the offline setting in which rich temporal context information is available. Taking this into consideration, the performance of our method is comparable to other offline setting methods. Compared to the baseline, the proposed method outperforms significantly, 0.06 higher mAP on average. Fig \ref{fig_result4} visualizes results in some challenging test data instances on ActivityNet.

\subsection{\textcolor{black}{Additional Analysis}}
\begin{table}
\caption{Per-frame labeling performance on THUMOS'14.}
\centering
\includegraphics[trim=11.4cm 8.6cm 11.4cm 8.45cm, clip=true, width=0.21\linewidth]{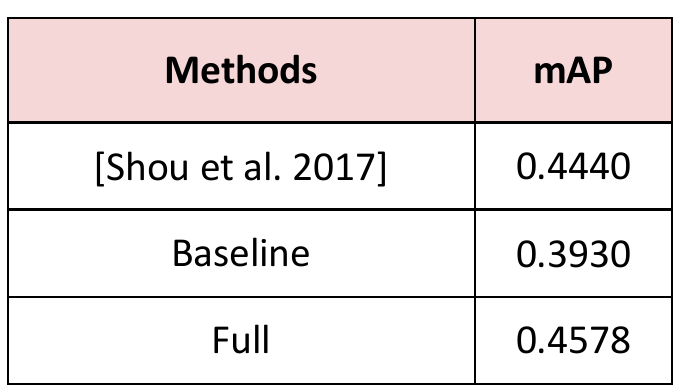}
\label{cmp_table_perframe}
\end{table}

\noindent\textbf{\textcolor{black}{Performance in online setting}}: \textcolor{black}{Since the proposed method works in the online setting, it is not straightforward to directly compare the performance with methods based on the offline setting. To our best knowledge, there is no metric for the online action detection performance. Thus, we evaluate the per-frame mAP performance as exploited by Shou et al. \cite{shou2017cdc}. Tab. \ref{cmp_table_perframe} shows the performance on THUMOS'14. The proposed method outperforms Shou et al. by 0.01.}

\noindent\textbf{\textcolor{black}{Computational complexity}}: \textcolor{black}{We tested the proposed framework on a single NVIDIA Titan X GPU with 12GB memory. The speed of the proposed framework is around 9 frames per second (fps). Assuming that each C3D network in the proposed framework has three convolutional layers streams to deal with different temporal resolutions, similar to multi-scale methods, the fps of this configuration decreases to 7 fps, which implies that the proposed data augmentation allows less computational cost and efficient memory usage.}

\subsection{\textcolor{black}{Ablation Study}}
\begin{table}
\caption{Performance of proposed components on THUMOS'14.}
\centering
\includegraphics[trim=5.4cm 6.8cm 5.4cm 7.65cm, clip=true, width=0.55\linewidth]{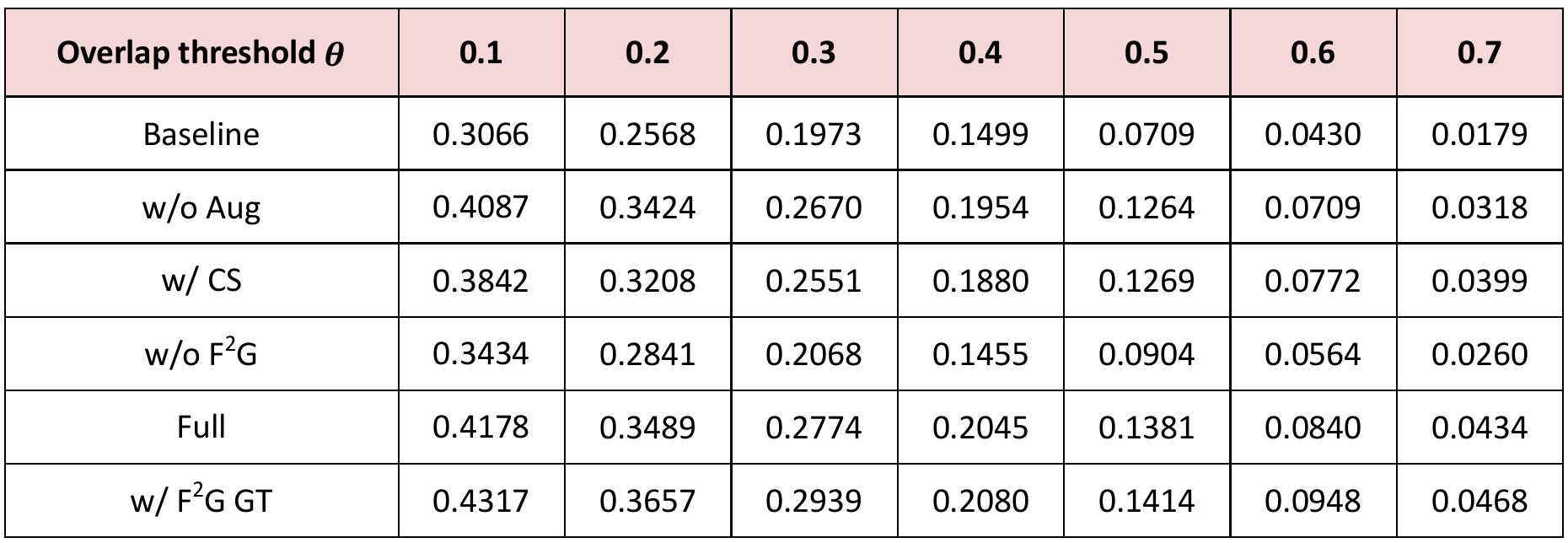}
\label{table_result3}
\end{table}

\begin{table}
\caption{Performance of proposed components on ActivityNet.}
\centering
\includegraphics[trim=9.4cm 6.8cm 9.4cm 7.65cm, clip=true, width=0.31\linewidth]{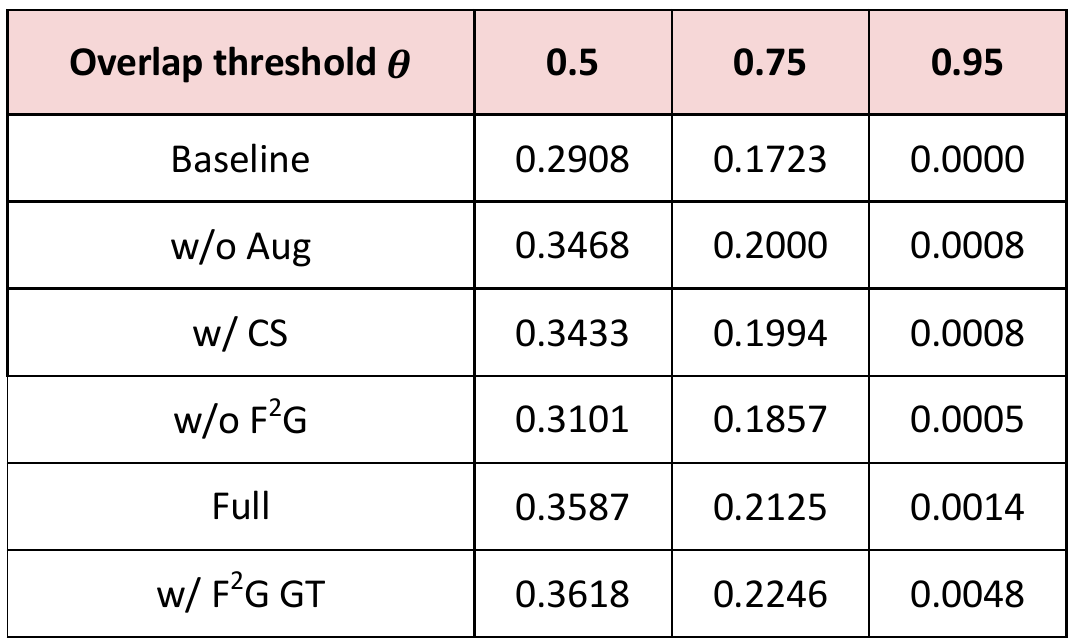}
\label{table_result3_act}
\end{table}

\textcolor{black}{We conduct additional experiments to analyze the impact of each model component by eliminating them one at a time. The experiments are conducted with six model setups: i) baseline, ii) without data augmentation (w/o Aug), iii) the C3D$_{AR}$ network connected after the C3D$_{PR}$ network (w/ CS), iv) without the F$^2$G network (w/o F$^2$G), v) the full model (Full), and vi) ground truth as future frame generation output (w/ F$^2$G GT). Tab. \ref{table_result3} summarizes the result on THUMOS'14 dataset and Tab. \ref{table_result3_act} the result on ActivityNet.}

\vspace{.3cm}
\noindent\textcolor{black}{\textbf{Baseline}: This setup is when the proposed framework consists of PR, AR, and Det networks. The performance is shown in the second row of Tab. \ref{table_result3} and Tab. \ref{table_result3_act}. Its performance is
inferior to all other settings.}

\vspace{.3cm}
\noindent\textcolor{black}{\textbf{Data augmentation}: Without data augmentation for model training, the performance decreases by 0.01 on average on both datasets (the third row). This result indicates that with proper data augmentation, including the frame interpolation used in the proposed framework, the performance can improve more.}

\vspace{.3cm}
\noindent\textcolor{black}{\textbf{Proposal representation and action representation configuration}: This setup is to study which of two C3D network arrangements, in parallel or serial, is useful in the proposed framework. As exploited in \cite{shou2016temporal}, the C3D$_{AR}$ network only takes the input segment classified as action by the C3D$_{PR}$ network. In this setup, the performance decreases, on average, by 0.02 and 0.01 on THUMOS'14 and ActivityNet,
respectively (the fourth row).}

\vspace{.3cm}
\noindent\textcolor{black}{\textbf{Future frame generation}: Exploiting future frame generation component increases the mAP, on average, by 0.05 on THUMOS'14 and 0.02 on ActivityNet, respectively (the fifth row). The performance gain is the most significant among other components. The result demonstrates that using the F$^2$G network allows our framework to consider more amount of information when compared to a situation without the F$^2$G. Thus, the limitations in the online setting is resolved to a certain extent.}

\vspace{.3cm}
\noindent\textcolor{black}{\textbf{Ground truth as future frame generation output}: To simulate this setup, we replace the output of the F$^2$G network by the ground truth frames. In other words, given an input sequence $(I_{t - 15}, ..., I_t)$ the F$^2$G generates future frames $(\hat{I}_{t + 1}, ..., \hat{I}_{t + 7})$; we replace generated future frames $(\hat{I}_{t + 1}, ..., \hat{I}_{t + 7})$ by the actual future frames $({I}_{t + 1}, ..., {I}_{t + 7})$ to simulate the aforementioned situation. The results are shown in the bottom row of Tab. \ref{table_result3} and Tab. \ref{table_result3_act}. Comparing to the performance of the `full' model  (the sixth row), using the ground truth as outputs of the F$^2$G network (the bottom row) increases the mAP, on average, by 0.01 on THUMOS'14 and 0.006 on ActivityNet. This result indicates that improving future frame generation performance leads to detection performance increase.}

\textcolor{black}{To summarize, as we argued in Sec. \ref{sec_intro}, for the online action detection scenario from video streams, a limited amount of information is a significant factor; using the F$^2$G network resolves this limitation by feeding predicted future input frames of a short period, eight frames in this paper, to the system so that more information is considered. Augmenting data also improves the detection performance, which means that making a model aware of variation of action duration is critical to a certain extent. Arranging two C3D networks in parallel, instead of connecting them in serial, is more effective in the proposed framework.}

\subsection{Limitations} 
We demonstrated that the proposed method shows comparable performances on two benchmark datasets. However, there are several limitations that we summarize as follows.

\begin{enumerate}
\item[--] Computational complexity: The proposed framework exploits four deep neural networks, which costs roughly 174M parameters. The reason is due to the difficulties of the online action localization task, which requires several components to deal with the lack of available information, distinguishing actions from background scenes, and accurately localize the start and the end of an action class. Thus, we designed the proposed framework with four deep neural networks in each of which is dedicated to deal with the issues mentioned above.

\item[--] Limited backpropagation during training: This limitation comes from the limited GPU resources to handle all four networks at the same time. As described in Sec. 3, each network of the proposed framework is trained separately, which implies that detection error at the final network LSTM$_{DET}$ does not backpropagate to the input layer of AR, PR, and F$^2$G networks, respectively. 

\item[--] Dependency on the F$^2$G network: We demonstrated that using generated future frames improves the online temporal action localization performance. However, the generation performance is not satisfactory when compared to the real ground truth frames. There is a large room for improving generation performance in which the proposed method depends.

\item[--] Room for further improvement: As we mentioned above, the limitations of the proposed method mostly come from the hardware side. We expect that, with enough computational resources, training the proposed framework with proper backpropagation, the performance will improve to a certain extent.
\end{enumerate}

\section{Conclusion and Future work}
\label{sec6}
In this paper, we proposed a novel action detection framework to address the challenging problems of online action detection from untrimmed video streams. To resolve the limited information issue, we proposed to exploit a future frame generation network. To learn temporal order using only visual information without learning any temporal prior, such as duration of action, we reorganized action class as two temporally-ordered subclasses. To make the proposed framework generalize better, we augmented training video data by varying the duration of action. 

We demonstrated that the performance of the proposed framework is comparable with the offline setting methods on two benchmark datasets, THUMOS'14 and ActivityNet. Through the ablation study, we demonstrated that the F$^2$G network gives meaningful improvement. We believe that other time-series tasks, such as traffic flow prediction \cite{POLSON20171} and financial market analysis \cite{CHONG2017187}, can also be benefitted by using a future generation network. In the meanwhile, there are also several limitations. The dependency on the future frame generation network and computational complexity of the proposed framework need to be addressed for further improvement. 

As future work, we plan to design a more efficient feature extraction network so that the whole framework can learn with the same backpropagation error. We will also plan to formulate action detection as a multitask learning problem.

\bibliographystyle{unsrt}  
\bibliography{template}  

\end{document}